\pdfoutput=1

\documentclass[11pt]{article}

\usepackage[]{acl}

\usepackage{times}
\usepackage{latexsym}

\usepackage[T1]{fontenc}

\usepackage[utf8]{inputenc}

\usepackage{microtype}

\usepackage{inconsolata}
\usepackage{graphicx}
\usepackage{tabularx}
\usepackage{soul}

\usepackage{multirow}
\usepackage{amsmath,amssymb,amsfonts}
\usepackage{amsthm}
\usepackage{mathrsfs}
\usepackage[title]{appendix}%
\usepackage{xcolor}%
\usepackage{textcomp}%
\usepackage{manyfoot}%
\usepackage{booktabs}%
\usepackage{algorithm}%
\usepackage{algorithmicx}%
\usepackage{algpseudocode}%
\usepackage{listings}%
\usepackage{float}
\usepackage{makecell}
\usepackage{ulem}

\definecolor{hengpink}{cmyk}{0, 0.7808, 0.4429, 0.1412}

\title{ProSwitch: Knowledge-Guided Instruction Tuning to Switch Between Professional and Non-Professional Responses}

\author{
Chang Zong$^1$, Yuyan Chen$^2$, Weiming Lu$^3$, Jian Shao$^3$ \\ \textbf{Yongfeng Huang$^4$, Heng Chang$^5$, Yueting Zhuang${^3}$\thanks{$^\dagger$Corresponding authors.}} \\ $^1$Zhejiang University of Science and Technology\\ $^2$Fudan University, $^3$Zhejiang University \\ $^4$The Chinese University of Hong Kong, $^5$Tsinghua University
}

\begin{document}
\maketitle
\begin{abstract}
Large Language Models (LLMs) have demonstrated efficacy in various linguistic applications, including question answering and controlled text generation. However, studies into their ability to switch between opposite styles of responses in professional domains remain underexplored. This study introduces a novel approach, named \textbf{ProSwitch}, which enables a language model to switch between professional and non-professional answers, by tuning and evaluating through the guidance of domain and style knowledge. ProSwitch unfolds in three phases: LLM-augmented preparation to collect domain knowledge and QA pairs, instruction tuning to optimize LLMs with multiple levels of knowledge, and comprehensive evaluation to assess both style discrimination and reference-based quality of the generated text. Comparative analysis of ProSwitch against general and specialized LLMs reveals that our approach outperforms baselines in switching between professional and non-professional responses.
\end{abstract}

\section{Introduction}
\begin{figure}[htbp]
    \begin{center}
    \includegraphics[width=0.48\textwidth]{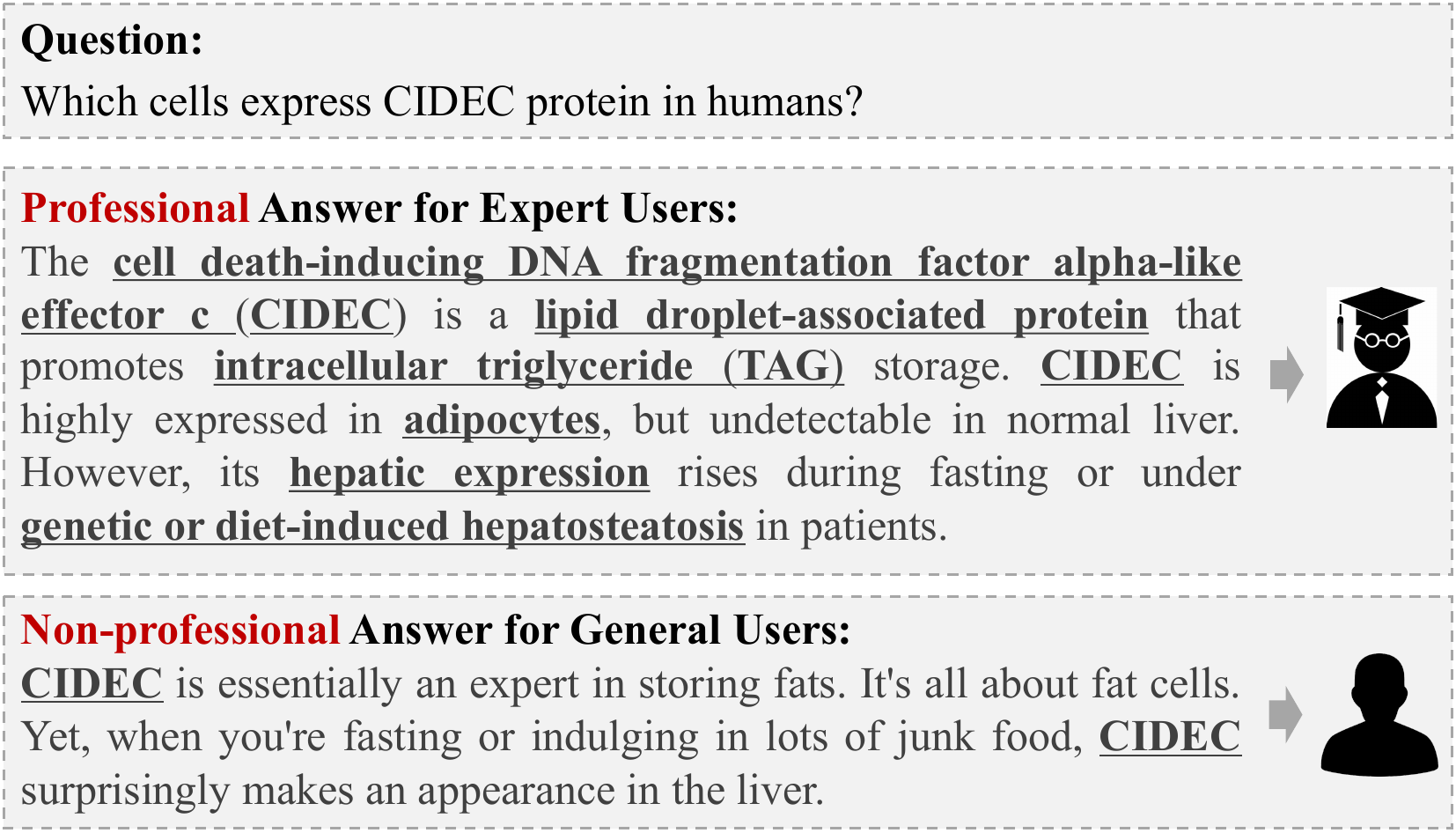} 
    \caption{Two answers cater to two different types of users, where the professional one contains more technical terminology and richer semantic information.}
    \label{example}
    \end{center}
\end{figure}

Providing text responses in various styles is an important task in professional fields, where users with different backgrounds can access the services they need. Figure \ref{example} shows a scenario for answering biomedical questions, where answers can be provided in both styles to improve the efficiency of information retrieval for different users. Meanwhile, Large Language Models (LLMs), such as ChatGPT and Llama \cite{llama}, have excelled in natural language tasks, including question answering \cite{zong-etal-2024-triad,omar2023chatgpt,tan2023chatgpt} and information extraction \cite{perot2023lmdx,PromptIE}. In specific domains, LLMs can provide answers that fit a particular style by integrating domain knowledge, as seen with ChatDoctor \cite{chatdoctor}, ChatLaw \cite{chatlaw} and FinGPT \cite{fingpt}. However, LLMs remain underutilized to address language style variation tasks in professional domains. 

Previous research aims to address language professionalism problems from various disciplinary perspectives. Some studies in linguistics and pedagogy focus on describing the characteristics of professional and colloquial language \cite{malyuga2021expressive,orrego2016reception}, stating that the distinctive feature of professional language is the terminological lexicon and the logical structure. Other studies in computer science achieve style transfer aimed at experts and layman users \cite{pu-demberg-2023,xu2022self} through controllable text generation, in which a prompt describing the desired style is used to make LLMs generate content that closely mimics real scenarios \cite{li2024pre,zhou2023controlled,NEURIPS2021_d0f5edad}. However, some issues remain unexplored in existing studies. First, in addition to lexical style transfer in general scenarios, there is a lack of research on enabling LLMs to master style switching abilities within professional domains while considering both lexical and structural language features. Second, besides assessing the stylistic characteristics of the generated results, it is also necessary to quantitatively evaluate the overall performance of LLMs in text generation and style switching.

\begin{figure*}[htbp]
    \begin{center}
    \includegraphics[width=1\textwidth]{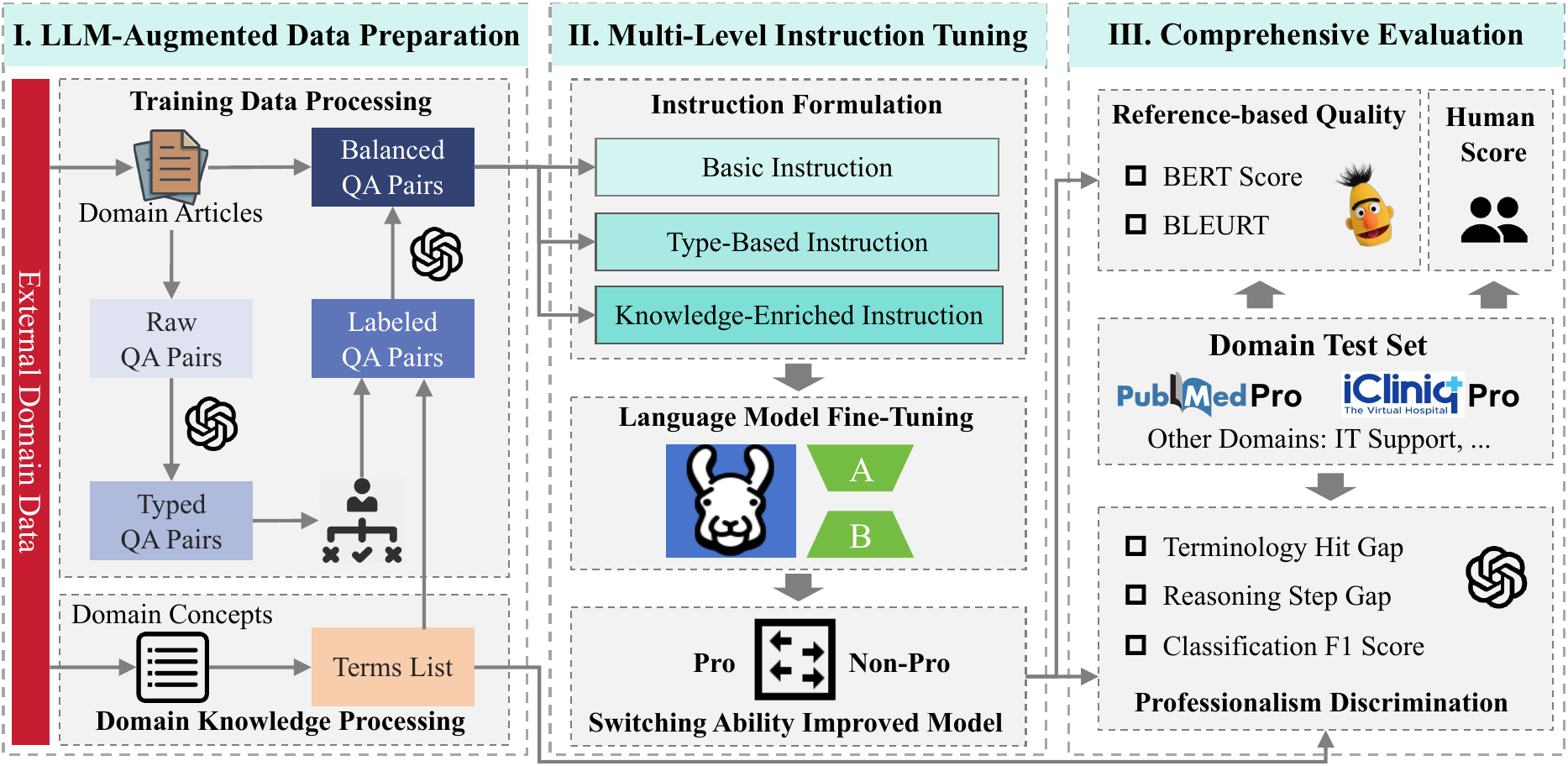} 
    \caption{Our ProSwitch method contains three phases to improve the style switching ability in professionalism, through exploiting domain knowledge for instruction tuning in multiple levels and performance evaluation.}
    \label{framework}
    \end{center}
\end{figure*}

To address the above issues, this study introduces a framework to improve and evaluate the ability of LLMs to switch between generating professional and non-professional answers, named \textbf{ProSwitch}. The process involves three stages, as shown in Figure \ref{framework}. In the data preparation phase, we generate a labeled and balanced QA dataset through a semi-automatic data augmentation process using domain-specific articles and concepts. To remedy the drawback of existing works regarding the first issue, we conduct instruction tuning by crafting multiple formulations of prompts to improve the style switching ability of an LLM by providing knowledge at different levels of granularity. To address the second issue, we propose a comprehensive evaluation strategy that contains indicators of both professionalism discrimination and reference-based language quality, fully leveraging the knowledge of professionalism proposed in previous research and the semantic understanding capabilities of LLMs. 

In summary, our contributions are as follows: 

\begin{itemize}
    \item We introduce \textbf{ProSwitch}, the first framework focusing on the generation of switching between professional and non-professional answers. Our framework exploits domain knowledge through instruction tuning, different from typical style transfer studies that focus only on lexical changes. 
    \item We propose and analyze instruction formulations from multiple levels to facilitate the tuning process by providing increasingly rich domain information, which is distinct from prompt-tuning and single-level instruction tuning used in previous work. 
    \item We perform a comprehensive evaluation by proposing indicators for both professionalism discrimination and language quality. Performance in QA datasets from the medical and IT professional domains reveals that ProSwtich outperforms both general and specialized LLMs.
\end{itemize}

\section{Related Work}
\subsection{Text Style Transfer Learning}
Text style transfer involves changing the style of an input sentence without altering its core meaning \cite{Jin_Jin_Hu_Vechtomova_Mihalcea_2022,Babakov_Dale_Logacheva_Panchenko_2022,Mir_Felbo_Obradovich_Rahwan_2019}. Previous studies have used sequence-to-sequence learning methods that apply parallel corpora with paired sentences in various styles \cite{xu2023specializing,Cheng_Gan_Zhang_Elachqar_Li_Liu_2020,hu-etal-2021}. However, due to the high demand for resources and costs for data labeling, parallel data in diverse styles is limited. This has encouraged a growing interest in investigating practical scenarios where only non-parallel stylized corpora are available \cite{jin2024reusing,Reif_Ippolito_Yuan_Coenen_Callison-Burch_Wei_2022,jin2024settp}.

\subsection{Controllable Text Generation}
Controllable text generation is a rapidly developing field dedicated to creating text or responses with designated characteristics \cite{Keskar_McCann_Varshney_Xiong_Socher_2019,Dathathri_Madotto_Lan_Hung_Frank_Molino_Yosinski_Liu_2019,He_Kryscinski_McCann_Rajani_Xiong_2021}. Various strategies have been suggested for this task, including sequence-to-sequence models that show potential in crafting excellent content tailored to particular needs. \cite{Wu_Liu_Liu_Stenetorp_Xiong_2021,Amplayo_Angelidis_Lapata_2021}. Other methods have also been introduced to improve text generation controllability, such as conditional generation \cite{He_Kryscinski_McCann_Rajani_Xiong_2021}, prompt-based generation \cite{yang-etal-2023}, and multitask learning \cite{gu-etal-2022}.

\subsection{LLM Instruction Fine-Tuning}
Instruction tuning combines the best aspects of pre-train fine-tuning and prompting approaches via supervised fine-tuning \cite{Wei_Bosma_Zhao_Guu_Yu_Lester_Du_Dai_Le_2021}. In this way, a model is trained to sequentially predict each token in the output, given the instruction and input \cite{InstructGPT,BLOOMZ,alpaca,Vicuna}. Some other domain language models apply instruction tuning methods to solve specific tasks or scenarios, such as information extraction \cite{InstructUIE}, sentiment analysis \cite{InstructSentiment}, medical dialogue \cite{chatdoctor}, and code generation \cite{WizardCoder}. To quickly adapt LLMs to downstream tasks, efficient fine-tuning techniques, such as addition-based \cite{Schick_2021}, specification-based \cite{BitFit}, and re-parameterization-based \cite{hu2022lora}, optimize a small fraction of parameters.

Despite the progress mentioned above, the ability of LLMs to switch between professional and non-professional responses has not been fully explored, particularly from the perspectives of domain and stylistic knowledge.

\section{Preliminaries}
\subsection{Professionalism Definition} \label{professionalism}
Unlike formality transfer and text simplification \cite{rao-tetreault-2018,maddela-etal-2021}, our study focuses on professionalism switch, where both style features and domain knowledge should be acquired by the model to fulfill the task. Referring to previous studies on linguistics and education \cite{malyuga2021expressive,orrego2016reception,malyuga2012professional}, the professionalism of a sentence is gauged by analyzing \textbf{domain terminology} and \textbf{logical structure}, necessitating the quantification of \textbf{terms} and \textbf{reasoning steps}. Then, we define the professionalism of a sentence as:
\begin{equation}
\begin{aligned}
Pro(O)=f_i(f_t(O,L_\mathcal{T}), f_{r}(O,\mathcal{M}))
\label{eq:eq0}
\end{aligned}
\end{equation}, where $f_t(\cdot)$ and $f_r(\cdot)$ are functions to calculate domain terms and reasoning steps from the output sentence $O$, $f_i(\cdot)$ is the function of integrating two indicators, $L_\mathcal{T}$ is the list of terms to be matched, $\mathcal{M}$ is the model for reasoning parsing. When $Pro(O)$ meets a specific condition, the sentence $O$ can be treated as a professional-styled text.

\subsection{Task Formulation}
We propose to improve the ability of an LLM to switch between professional and non-professional styles, aiming to maximize the distinction between the text generated in two styles while maintaining the quality of generated sentences, by assessing with a set of detailed indicators. Our objective can be formulated as:
\begin{equation}
\begin{aligned}
&\max \big(f_p(O_{p}, O_{np}) + f_q(O_{p}) + f_q(O_{np})\big), \\
&O_{p} = LM(Pmt_{p}), O_{np} = LM(Pmt_{np})
\label{eq:eq1}
\end{aligned}
\end{equation}, where $m$ is the desired method to maximize the score of text generated by an LLM. $f_p(\cdot)$ and $f_q(\cdot)$ are evaluation functions to calculate the professionalism discrimination and the general quality of the generated text, respectively. $O_p$ and $O_{np}$ are outputs generated by language model $LM$, which is provided with prompts for professional style $Pmt_p$ and non-professional style $Pmt_{np}$.

\subsection{Prompt Formulation}
A prompt to generate answers by an LLM in a particular style can be regarded as a concatenation of three components: task and style guidelines, questions to be addressed, and LLM-related limit information for output consistency. The prompt used in our study can be formulated as:
\begin{equation}
\begin{aligned}
Pmt_p =& Guide_p \oplus Q_n \oplus Limit_{lm}, \\
Pmt_{np} =& Guide_{np} \oplus Q_n \oplus Limit_{lm}
\label{eq:eq2}
\end{aligned}
\end{equation}, where $Guide_p$ and $Guide_{np}$ are guidelines for generating professional and non-professional style answers. $Q_n$ is the $n$-th question that needs to be answered. $Limit_{lm}$ is the restrictive text for a specific language model $lm$. These components are connected with the concatenation operator $\oplus$. 

\section{Proposed ProSwitch}
\subsection{LLM-Augmented Data Preparation} \label{data-prepare}
\paragraph{Academic QA Pairs Collection.} Textual professional styles are often reflected in academic scenarios, particularly in knowledge-intensive fields such as healthcare and medicine. Meanwhile, professional-style features can be learned from specialized QA tasks. With the information above, we collected two medical QA datasets, BioASQ \cite{BioASQ} and PubMedQA \cite{PubMedQA}, sourced from academic articles. The responses in these datasets aim to clarify the questions based on a section of related papers, which are rich in technical terms and detailed explanations. We consider these datasets as the seeds of our professional-style training data.

\paragraph{Question Type Classification.} We have observed apparent style variations among different types of QA pairs. For instance, an answer using a list of terms differs significantly from an answer explaining a phenomenon with only words. This inspires us to categorize QA pairs by their question types to help a model learn the type-related features of professionalism. According to BioASQ, we consider four types: list, summarize, yes/no, and factoid. However, PubMedQA does not specify any types, so we employ GPT-4 to classify each QA pair into one of the four types by providing a few examples, followed by a manual check (details in Appendix \ref{typecls}). This LLM-supported type classification task can be formulated as :
\begin{equation}
\begin{aligned}
T(Q_n) & = LM(Pmt_t, (Q_n, A_n), L_t, \{S_1, ..., S_k\}) \\
L_t & = \{list, summarize, yes/no, factoid\}
\label{eq:eq3}
\end{aligned}
\end{equation}, where $Q_n$ and $A_n$ are the questions and answers that need to be classified. $Pmt_t$ is the prompt for the type classification task with label set $L_t$. ${S_1, ..., S_k}$ are the $k$ examples for performing a few-shot learning.

\paragraph{Data Balanced Augmentation.} Due to the lack of corresponding non-professional responses in our dataset and a shortage of QA pairs for training in both styles, there is a need to perform data augmentation for training. Using LLM and in-context learning (ICL) \cite{ICLSurvey}, our goal is to increasingly generate QA pairs for each question type in each style, striving for an adequate and equal size. GPT-4 is assigned to generate answers using either professional or non-professional language, adhering to specific guidelines based on the provided questions and examples. For professional data augmentation, GPT-4 is used exclusively to rephrase the referenced answers. For non-professional data generation, GPT-4 directly provides an answer in casual language, complying with the guidelines (details in Appendix \ref{dataaugment}). This data augmentation task can be formulated as follows:
\begin{equation}
\begin{aligned}
A(Q_n) =& LM(Pmt_a, Q_n, \{S_1, ..., S_k\}),\\
Pmt_a =& f_i(Dict, L_{p}, T(Q_n))
\label{eq:eq4}
\end{aligned}
\end{equation}, where $Pmt_a$ is the prompt for answering questions based on question types and style labels, which is retrieved from a pre-defined dictionary $Dict$ by an indexing function $f_i$, using the type of the question $T(Q_n)$ and the label $L_{p}$ as the keys.

\paragraph{Term Knowledge Processing.} Evaluating the discrimination of professionalism in our task requires domain-specific expertise, and domain terms help us perform the evaluation automatically. In the medical field, we gather MeSH \footnote{https://www.nlm.nih.gov/databases/download/mesh.html}, a widely utilized XML-formatted list of medical terms. We derive all \textit{QualifierNames} from the original file to compile a medical terminology list. The list is used as our external knowledge to match phrases in an answer to quantitatively evaluate its professionalism.

\subsection{Instruction Formulation} \label{instruction}
\subsubsection{Multi-Level Instruction} 
With the QA pairs generated in both professional and non-professional styles, we have to provide additional guidance to clarify the task for the language model during fine-tuning. Adhering to the Alpaca \cite{alpaca} instruction format, we further formulate instructions focusing on three levels of information for the style switching task, presented as follows.

\paragraph{Basic instruction.} Firstly, we only apply the superficial description of professional and non-professional styles. The instruction for professional answers is conveyed as: \textit{Answer the question and explain the reason with detailed steps using technical professional expressions.} For non-professional answers, the instruction is: \textit{Answer the question and explain the reason with a simple explanation using casual non-professional expressions.}

\paragraph{Type-based instruction.} In contrast, taking into account the significant differences in responses to various question types, we suggest a fine-grained instruction format by providing type-based descriptions such as applying \textit{Answer the question with \underline{a list of items} and explain \underline{each item} with...} for the list-type questions. This formulation results in a permutation of two style labels (professional and non-professional) and four question types (list, summary, yes/no, and factoid). 

\paragraph{Knowledge-enriched instruction.} Furthermore, with the rich expression information contained in domain-related articles, we propose a knowledge-enriched instruction by injecting question-related article snippets, which are treated as implicit knowledge of professional style, to construct professional instructions, formatted as: \textit{\underline{Knowledge: <article\_snippet>}. Answer the question following the style of the knowledge provided and ...}. For nonprofessional instructions, we inject a more descriptive sentence as explicit knowledge of non-professional style to explain what the answer should be expressed, formatted as: \textit{Knowledge: A non-professional answer is prone to use analogies and phrasal verbs to explain the question with fewer technological and organizational expressions. Answer the question following the knowledge using non-professional expressions}.

\subsubsection{LLM-Related Restrictive Information} 
Different language models have varying capabilities and can generate texts in distinctive lengths and formats, leading to inconsistent comparisons. To address this problem, during our testing phase, we add brief restrictive information as described in Equation \ref{eq:eq2} to the input questions, guiding language models to generate answers in a similar format. Specifically, since some models fine-tuned with human chat data tend to provide dialogue-style responses, we append \textit{Answer the question with a single paragraph containing detailed explanations.} to questions to avoid unrelated information and dissimilar formats.

\subsection{Automatic Evaluation}
\subsubsection{Professionalism Discrimination Scores}
To evaluate the ability of ProSwitch, we propose a set of indicators to demonstrate the discrimination between professional and non-professional styles of the generated output. Referring to the professionalism defined in Section \ref{professionalism}, we describe our indicators as follows. 

\paragraph{Terminology Hit Gap (THG).} The number of technical terms contained in a generated paragraph is a useful metric that leads us to introduce our first indicator. THG measures the disparity between the number of technical terms found in professional and non-professional responses. With the term knowledge collected, we compute this indicator by performing a phrase-level matching between the output of the language model and phrases in our domain term list, noted as:
\begin{equation}
\begin{aligned}
&THG = |\frac{1}{N} \sum_{n=1}^{N} TH_n^p - \frac{1}{N} \sum_{n=1}^{N} TH_n^{np}| , \\
&TH_n^p = f_c(f_m(Term_d, LM(Pmt_n^p)))
\label{eq:eq5}
\end{aligned}
\end{equation}, where $TH_n^p$ and $TH_n^{np}$ are the terminology hit values of the $n$-th answer in professional and non-professional styles, respectively. $f_m$ and $f_c$ are the functions for term matching and hit counting, respectively. $Term_d$ is the terminology list in domain $d$. $LM(Pmt_n^p)$ is the output generated by $LM$ with a prompt describing the $n$-th question professionally.

\paragraph{Reasoning Step Gap (RSG).} Furthermore, we propose our second indicator to distinguish the structure of generated sentences, RSG, which measures the gap in reasoning steps between professional and non-professional responses. This indicator is based on the notion that professional language typically exhibits a more rigorous logical structure than casual language. To calculate RSG, we use GPT-4 to transform the answer into sequential reasoning steps and then count the steps with a parsing function. This process can be noted below.
\begin{equation}
\begin{aligned}
RSG =& |\frac{1}{N} \sum_{n=1}^{N} RS_n^p - \frac{1}{N} \sum_{n=1}^{N} RS_n^{np} |, \\
RS_n^p =& f_p(LLM(Pmt_r, LM(Pmt_n^p)))
\label{eq:eq6}
\end{aligned}
\end{equation}, where $RS_n^p$ and $RS_n^{np}$ are the numbers of reasoning steps for the $n$-th professional and non-professional answers. $f_p$ is the parsing function to count steps from the reasoning details generated by an $LLM$. $Pmt_r$ is the prompt for GPT-4 to provide reasoning process from a given answer.

\paragraph{Pro F1.} To further measure the ability of our fine-tuned language model to generate desired text styles, we implement a binary classification task. We calculate the number of terms and reasoning steps contained in answers, and set thresholds for these two indicators based on their true labels (details in Appendix \ref{threshold}). For newly generated answers, we compare the labels satisfied by their indicators with their inherent labels to obtain the typical F1 score, denoted as Pro F1.

\subsubsection{Reference-based Scores}
To investigate whether our tuning stage degenerates the fundamental ability of an LLM, we employ BERT score \cite{BERTScore} and BLEURT \cite{sellam-etal-2020}, two reference-based machine learning metrics for text generation, which are able to capture semantic similarities between sentences using BERT models\cite{vaswani2017attention}. These metrics are illustrated as follows:
\begin{equation}
\begin{aligned}
BERT score = 2 \frac{P_{BERT} \cdot R_{BERT}}{P_{BERT}+R_{BERT}},
\label{eq:eq7}
\end{aligned}
\end{equation} where $P_{BERT}$ and $R_{BERT}$ are the precision and recall that are calculated with the contextualized word embeddings for the reference answer and the generated output of the $n$-th question, respectively.
\begin{equation}
\begin{aligned}
BLEURT &= Wv_{[CLS]}+b, \\
v_{[CLS]},v_{x_1},...,v_{x_r}&,v_{\tilde{x}_1},...,v_{\tilde{x}_p} = BERT(x,\tilde{x}),
\label{eq:eq8}
\end{aligned}
\end{equation} where $x_1, .., x_r$ to be the reference sentence of length $r$ and $\tilde{x}_1, .., \tilde{x}_p$ be a prediction sentence of length $p$, $v_{[CLS]}$ is the representation for the special $[CLS]$ token, $W$ and $b$ are the weight matrix and bias vector respectively.

\section{Evaluation and Analysis}

\subsection{Dataset}
We develop two domain datasets, \textbf{PubMedPro} and \textbf{IcliniqPro}, to assess the style switching ability. PubMedPro, which is constructed following the Alpaca format as detailed in Section \ref{data-prepare}, comprises 24,000 QA pairs in both professional and non-professional styles within the medical field. We select 200 questions of different types, with their corresponding answers in positive and negative styles, as our test set for evaluation. These questions originate from BioASQ \cite{BioASQ} and PubMedQA \cite{PubMedQA}, two freely accessible QA datasets drawn from PubMed's academic articles\footnote{https://pubmed.ncbi.nlm.nih.gov/}. Another dataset is IcliniqPro, derived from iCliniq\footnote{https://www.icliniq.com/qa/medical-conditions}, a medical dialogue dataset downloaded from the repositories mentioned in \cite{zeng-etal-2020,MQS-ECL}. We manually and carefully select questions with the same number and similar expressions as those in PubMedPro, according to two principles: 1. The questions need to be answered with specific knowledge; 2. The questions are stated directly without personal feelings.
 
\subsection{Baselines}
We evaluate ProSwitch variants against multiple baselines. Llama2-Chat \cite{llama2}, our foundation model, is a prevalent language model for general dialogue scenarios. ChatDoctor \cite{chatdoctor} is a specialized language model fine-tuned with extensive patient-doctor dialogue data to improve the accuracy of medical advice. GPT-4-Turbo \footnote{https://platform.openai.com/docs/models/gpt-4} is a commercial LLM for general purposes with a much larger parameter size. Furthermore, we implement the work on formality style transfer (FST) \cite{etinger2019formality} as a traditional baseline. We reproduce the expert-layman study \cite{pu-demberg-2023} as a competitor in controllable text summarization (CTS).

\subsection{Instruction Tuning}
With the prompts that contain multiple levels of instructions and restrictive information, as described in Section \ref{instruction}, we apply LoRA \cite{hu2022lora}, one of the parameter-efficient fine-tuning methods (PEFT), to optimize a pre-trained model for our task. LoRA significantly minimizes the number of trainable parameters by freezing the pre-trained model weights and incorporating trainable rank decomposition matrices into the Transformer. Finally, using Llama2-Chat as our base model, three optimized models are generated according to the three levels of instruction formulations, named \textbf{ProSwitch-B}, \textbf{ProSwitch-T}, and \textbf{ProSwitch-K}, respectively.

\begin{table*}[htbp]
\centering
\begin{tabular}{l|cccccccc}
\bottomrule
\multirow{3}{*}{\textbf{Models}} & \multicolumn{5}{c}{\textbf{PubMedPro}} & \multicolumn{3}{c}{\textbf{IcliniqPro}} 
\\ &
\multicolumn{3}{c}{\textbf{Style Professionalism}} & \multicolumn{2}{c}{\textbf{Reference-based Quality}} & \multicolumn{3}{c}{\textbf{Style Professionalism}}
\\  & 
\multicolumn{1}{c}{\textbf{THG}} & \multicolumn{1}{c}{\textbf{RSG}} & \multicolumn{1}{c}{\textbf{Pro F1}} & \multicolumn{1}{c}{\textbf{BERT Score}} & \textbf{BLEURT} & \multicolumn{1}{c}{\textbf{THG}} & \multicolumn{1}{c}{\textbf{RSG}} & \multicolumn{1}{c}{\textbf{Pro F1}}
\\ \hline      
\multirow{1}{*}{Llama2-Chat} & \multicolumn{1}{c}{2.92} & \multicolumn{1}{c}{0.58} & \multicolumn{1}{c}{0.63} & \multicolumn{1}{c}{0.7292} & \multicolumn{1}{c}{0.4852} & \multicolumn{1}{c}{2.28} & \multicolumn{1}{c}{0.62} & \multicolumn{1}{c}{0.51}
\\
\multirow{1}{*}{ChatDoctor} & \multicolumn{1}{c}{1.74} & \multicolumn{1}{c}{0.33} & \multicolumn{1}{c}{0.60} & \multicolumn{1}{c}{0.7204} & \multicolumn{1}{c}{0.5012} & \multicolumn{1}{c}{1.68} & \multicolumn{1}{c}{0.89} & \multicolumn{1}{c}{0.44}
\\
\multirow{1}{*}{GPT-4-Turbo} & \multicolumn{1}{c}{3.19} & \multicolumn{1}{c}{0.63} & \multicolumn{1}{c}{0.66} & \multicolumn{1}{c}{0.7595} & \multicolumn{1}{c}{\underline{0.5484}} & \multicolumn{1}{c}{2.79} & \multicolumn{1}{c}{1.56} & \multicolumn{1}{c}{0.70}
\\
\multirow{1}{*}{FST} & \multicolumn{1}{c}{0.48} & \multicolumn{1}{c}{0.46} & \multicolumn{1}{c}{0.62} & \multicolumn{1}{c}{0.6948} & \multicolumn{1}{c}{0.4121} & \multicolumn{1}{c}{-} & \multicolumn{1}{c}{-} & \multicolumn{1}{c}{-}
\\
\multirow{1}{*}{CTS} & \multicolumn{1}{c}{1.68} & \multicolumn{1}{c}{0.84} & \multicolumn{1}{c}{0.65} & \multicolumn{1}{c}{0.7322} & \multicolumn{1}{c}{0.5442} & \multicolumn{1}{c}{1.32} & \multicolumn{1}{c}{2.57} & \multicolumn{1}{c}{0.69}
\\
\multirow{1}{*}{ProSwitch-B} & \multicolumn{1}{c}{\underline{3.44}} & \multicolumn{1}{c}{0.74} & \multicolumn{1}{c}{0.70} & \multicolumn{1}{c}{0.7472} & \multicolumn{1}{c}{0.5268} & \multicolumn{1}{c}{\underline{3.38}} & \multicolumn{1}{c}{3.04} & \multicolumn{1}{c}{0.76}
\\
\multirow{1}{*}{ProSwitch-T} & \multicolumn{1}{c}{\textbf{4.04}} & \multicolumn{1}{c}{\underline{1.06}} & \multicolumn{1}{c}{\underline{0.73}} & \multicolumn{1}{c}{\underline{0.7676}} & \multicolumn{1}{c}{0.5385} & \multicolumn{1}{c}{\textbf{3.58}} & \multicolumn{1}{c}{\underline{3.31}} & \multicolumn{1}{c}{\textbf{0.81}}
\\
\multirow{1}{*}{ProSwitch-K} & \multicolumn{1}{c}{3.26} & \multicolumn{1}{c}{\textbf{2.32}} & \multicolumn{1}{c}{\textbf{0.77}} & \multicolumn{1}{c}{\textbf{0.7799}} & \multicolumn{1}{c}{\textbf{0.5479}} & \multicolumn{1}{c}{3.30} & \multicolumn{1}{c}{\textbf{3.84}} & \multicolumn{1}{c}{\underline{0.79}}
\\ \bottomrule
\end{tabular}
\caption{The performance of three ProSwitch variants using three levels of instruction formulations, against LLMs, style transfer model (FST), and controllable text summarization method (CTS) on two datasets. THG and RSG are our proposed professionalism discrimination indicators. Pro F1 is the F1 score of the stylistic binary classification. We only record professionalism indicators for IcliniqPro dataset as the absence of ground-truth professional answers as references. The optimal and suboptimal scores are highlighted with bold and underlined text, respectively.}
\label{performance1}
\end{table*}

\subsection{Implementation Details}
During the tuning phase, we train our ProSwitch model on 24,000 QA pairs, evenly distributed between two style labels and four question types. This training process, conducted on an NVIDIA RTX A6000 GPU for three epochs, has a learning rate of 2e-5 and a batch size of 128, taking roughly four hours in total. For evaluation, the thresholds for terminology hit count and reasoning step count are set at 1 and 4 respectively, aligning closely with human labeling results with an AUC greater than 0.85. We evaluate the ChatDoctor baseline with 13B parameters and Llama2-Chat with 7B parameters, identical to our ProSwitch. All experiments are conducted on average three times generation.

\subsection{Professional Style Switching Performance}
We evaluate ProSwitch and the baselines on the PubMedPro dataset using both professionalism discrimination and reference-based quality metrics. Meanwhile, we evaluate models on the IcliniqPro dataset, focusing solely on professionalism discrimination due to the absence of professional ground-truth answers. The experimental results are shown in Table \ref{performance1}. We can observe from the results that:

\paragraph{ProSwitch outperforms baselines on all datasets.} Our knowledge-guided instruction tuning procedure improves the ability to switch between professional and non-professional styles, without compromising text generation capabilities.
\paragraph{Fine-grained instructions capture features better.} The model that uses type-based and knowledge-enriched instructions can provide more detailed guidance, leading to the generation of text in styles with higher expectations.
\paragraph{Specialized models downgrade switching ability.} ChatDoctor and FST, though fine-tuned with domain data or targeted at similar tasks, suffer a loss of style switching, which appears already partially equipped by other baselines.
\paragraph{Performance deficiency remains on large models.} GPT-4-Turbo and CTS, though trained with enormous corpora and tuned with numerous tasks, still show deficiency on the style switching task compared to our task-specific method.

A discussion on the variation of answers generated by LLMs is in Appendix \ref{variation}. A study on some representative cases is in Appendix \ref{casestudy}.

\subsection{Human Evaluation}
We recruit volunteers from different research backgrounds to evaluate a crowd-sourcing task. Each answer in the test set is rated in three aspects: style discrimination (SD), language fluency (LF), and answer accuracy (Acc). For style discrimination, each pair of answers is scored 1–5 points to determine the degree to which the two answers can be distinguished in professionalism. For fluency, each answer is also scored 1–5. For accuracy, each answer is individually labeled as true, false, or unknown by medical experts. Following \cite{xu2022self}, we calculate the percentage of ratings with 4 and 5 points as success rates (SR), and also the average scores (AS) (details in Appendix \ref{humaneval}). Human evaluation is performed on all baseline models and ProSwitch variants. The results of the human evaluation shown in Table \ref{human} show that our method improves style differentiation while maintaining fluency and accuracy.
\begin{table}[htbp]
\centering
\begin{tabular}{l|ccccc}
\bottomrule
\multirow{2}{*}{\textbf{Models}} & \multicolumn{2}{c}{\textbf{SD}} & \multicolumn{2}{c}{\textbf{LF}}  & \multicolumn{1}{c}{\textbf{Acc}} 
\\ &
\multicolumn{1}{c}{\textbf{AS}} & \multicolumn{1}{c}{\textbf{SR}} & \multicolumn{1}{c}{\textbf{AS}} & \multicolumn{1}{c}{\textbf{SR}} & \multicolumn{1}{c}{-}
\\ \hline      
\multirow{1}{*}{Llama2-Chat} & \multicolumn{1}{c}{3.60} & \multicolumn{1}{c}{0.57} & \multicolumn{1}{c}{3.97} & \multicolumn{1}{c}{0.78} & \multicolumn{1}{c}{0.65}
\\
\multirow{1}{*}{ChatDoctor} & \multicolumn{1}{c}{3.02} & \multicolumn{1}{c}{0.38} & \multicolumn{1}{c}{\underline{4.27}} & \multicolumn{1}{c}{\textbf{1.00}} & \multicolumn{1}{c}{\textbf{0.75}}
\\
\multirow{1}{*}{GPT-4-Turbo} & \multicolumn{1}{c}{3.75} & \multicolumn{1}{c}{0.82} & \multicolumn{1}{c}{\textbf{4.43}} & \multicolumn{1}{c}{\textbf{1.00}} & \multicolumn{1}{c}{\underline{0.72}}
\\
\multirow{1}{*}{FST} & \multicolumn{1}{c}{2.38} & \multicolumn{1}{c}{0.08} & \multicolumn{1}{c}{3.32} & \multicolumn{1}{c}{0.51} & \multicolumn{1}{c}{0.56}
\\
\multirow{1}{*}{CTS} & \multicolumn{1}{c}{3.18} & \multicolumn{1}{c}{0.47} & \multicolumn{1}{c}{4.11} & \multicolumn{1}{c}{\underline{0.97}} & \multicolumn{1}{c}{0.69}
\\
\multirow{1}{*}{ProSwitch-B} & \multicolumn{1}{c}{3.91} & \multicolumn{1}{c}{0.78} & \multicolumn{1}{c}{4.05} & \multicolumn{1}{c}{0.92} & \multicolumn{1}{c}{0.67}
\\
\multirow{1}{*}{ProSwitch-T} & \multicolumn{1}{c}{\textbf{4.28}} & \multicolumn{1}{c}{\underline{0.90}} & \multicolumn{1}{c}{4.13} & \multicolumn{1}{c}{\textbf{1.00}} & \multicolumn{1}{c}{0.67}
\\
\multirow{1}{*}{ProSwitch-K} & \multicolumn{1}{c}{\underline{4.23}} & \multicolumn{1}{c}{\textbf{0.93}} & \multicolumn{1}{c}{4.25} & \multicolumn{1}{c}{\textbf{1.00}} & \multicolumn{1}{c}{\underline{0.72}}
\\ \bottomrule
\end{tabular}
\caption{Human evaluation results. ProSwitch models are confirmed to generate answers with more professionalism discrimination (SD) while maintaining fluency (LF) and accuracy (Acc).}
\label{human}
\end{table}

\subsection{Effect of Tuning Strategy}
Apart from tuning language models using PEFT methods, we also attempt to train a ProSwitch model with a full parameter fine-tuning approach, to investigate the potential capacity of a foundation language model to learn how to switch between professional and non-professional styles. The performance of ProSwitch using LoRA and full fine-tuning (FFT) methods is shown in Table \ref{performance2}. We discover that the fully fine-tuned model tends to generate long answers with more reasoning steps, but with fewer technical terms, leading to a lower THG score, which indicates that fully fine-tuning learns expression better than wording.

\begin{table}[htbp]
\centering
\begin{tabular}{l|ccc}
\bottomrule
\multirow{1}{*}{\textbf{Models}}  & \multicolumn{1}{c}{\textbf{THG}} & \multicolumn{1}{c}{\textbf{RSG}} & \multicolumn{1}{c}{\textbf{Pro F1}}
\\ \hline      
\multirow{1}{*}{ProSwitch-Avg (LoRA)} & \multicolumn{1}{c}{3.58} & \multicolumn{1}{c}{1.37} & \multicolumn{1}{c}{0.73}
\\
\multirow{1}{*}{ProSwitch-Avg (FFT)} & \multicolumn{1}{c}{3.35} & \multicolumn{1}{c}{1.60} & \multicolumn{1}{c}{0.77}
\\ \bottomrule
\end{tabular}
\caption{The average value of the model (ProSwitch-Avg) trained on three types of instruction formulations with LoRA and full fine-tuning strategies in terms of professionalism discrimination indicators.}
\label{performance2}
\end{table}

\subsection{Logical Density Analysis}
As professional language is generally treated as a logically structured system \cite{malyuga2012professional}, answers with more reasoning text are considered more professional, which can also lead to longer text. However, our statistical results in Table \ref{performance3} show that answers that contain many logically linked concise sentences can fit the professional style well. By contrast, the density of reasoning steps within an answer is an effective feature of professionalism.

\begin{table}[htbp]
\centering
\begin{tabular}{l|ccc}
\bottomrule
\multirow{1}{*}{\textbf{Models}}  & \multicolumn{1}{c}{\textbf{Avg.Len}} & \multicolumn{1}{c}{\textbf{Avg.RS}} & \multicolumn{1}{c}{\textbf{RD}}
\\ \hline      
\multirow{1}{*}{Llama2-Chat} & \multicolumn{1}{c}{418.5} & \multicolumn{1}{c}{5.29} & \multicolumn{1}{c}{0.013}
\\
\multirow{1}{*}{ChatDoctor} & \multicolumn{1}{c}{443.5} & \multicolumn{1}{c}{5.83} & \multicolumn{1}{c}{0.013}
\\
\multirow{1}{*}{GPT-4-Turbo} & \multicolumn{1}{c}{724.0} & \multicolumn{1}{c}{7.05} & \multicolumn{1}{c}{0.010}
\\
\multirow{1}{*}{ProSwitch-Avg} & \multicolumn{1}{c}{336.0} & \multicolumn{1}{c}{5.92} & \multicolumn{1}{c}{\textbf{0.018}}
\\ \bottomrule
\end{tabular}
\caption{The average answer length (Avg.Len), average reasoning steps (Avg.RS), and reasoning density (RD) of the professional answers generated by different models. The higher reasoning density of ProSwitch indicates the more professional text it generates.}
\label{performance3}
\end{table}

\subsection{Domain Adaptation Analysis}
We evaluate the adaptation performance of ProSwitch in another domain: IT technical support, using the TechQA dataset \cite{castelli-etal-2020}, which contains real-world questions posed by users on the IBMDeveloper forum. This domain offers multifaceted professional support to accommodate users with diverse needs. We randomly select a subset of TechQA IT questions as the test set and collect 352 IT terms and their synonyms from various websites to calculate the THG indicator. We directly employ the ProSwitch models that were trained with medical QA datasets to generate professional and non-professional answers for IT questions. The performance compared to general models such as Llama2-Chat and GPT-4-Turbo is shown in Table \ref{domain}.

\begin{table}[htbp]
\centering
\begin{tabular}{l|ccc}
\bottomrule
\multirow{1}{*}{\textbf{Models}}  & \multicolumn{1}{c}{\textbf{Pro F1}} & \multicolumn{1}{c}{\textbf{THG}} & \multicolumn{1}{c}{\textbf{RSG}}
\\ \hline      
\multirow{1}{*}{Llama2-Chat} & \multicolumn{1}{c}{0.46} & \multicolumn{1}{c}{0.18} & \multicolumn{1}{c}{0.83}
\\
\multirow{1}{*}{GPT-4-Turbo} & \multicolumn{1}{c}{0.51} & \multicolumn{1}{c}{0.19} & \multicolumn{1}{c}{1.94}
\\
\multirow{1}{*}{ProSwitch-B} & \multicolumn{1}{c}{0.57} & \multicolumn{1}{c}{0.24} & \multicolumn{1}{c}{2.85}
\\
\multirow{1}{*}{ProSwitch-T} & \multicolumn{1}{c}{0.63} & \multicolumn{1}{c}{0.29} & \multicolumn{1}{c}{{3.02}}
\\
\multirow{1}{*}{ProSwitch-K} & \multicolumn{1}{c}{\textbf{0.69}} & \multicolumn{1}{c}{\textbf{0.36}} & \multicolumn{1}{c}{\textbf{4.07}}
\\ \bottomrule
\end{tabular}
\caption{The performance of ProSwitch variants on answering technical questions from IT domain without training with domain datasets. Our method consistently generates answers with more professional and non-professional features.}
\label{domain}
\end{table}

\section{Conclusion}
This study proposes ProSwitch, a knowledge-guided instruction tuning method, to improve the ability of language models to switch between professional and non-professional responses. We focus on domain question-answering tasks and conduct three phases: LLM-augmented data preparation, knowledge-guided multi-level instruction tuning, and comprehensive evaluation to acquire the ability of style discrimination and language generation quality in professional domains. Our findings demonstrate that ProSwitch significantly improves the style differentiation of generated text, compared to both general LLMs and specialized methods. 

\section*{Limitations}
The limitation of our research lies in the following three aspects. First, due to a lack of research that can be referenced as the formal definition of language professionalism, we have to propose our definition of professionalism and style discrimination quantitatively, which needs to be further verified and explored. Second, the deficiency of ground-truth datasets with clearly professional and non-professional labels forces our experiments to be conducted with massive generative data, which have not been fully inspected in terms of quality. Third, our method still awaits verifications in more specialized domains and with various foundation models. Our future studies will focus on addressing the above limitations.

\section*{Ethics Considerations}
All datasets utilized in this study are publicly available and we have adhered to ethical considerations by not introducing additional information as input during LLM training and LLM text generation.


\bibliography{anthology,custom}

\appendix
\section{LLM Prompts}
\subsection{Prompts for Type Classification}
We perform an LLM-based question type classification task by providing the following prompt in Table \ref{prompt1} to GPT-4 and replacing the <question> variable with our real questions in our datasets.
\begin{table}[H]
\centering
\begin{tabularx}{8cm}{|X|}
\hline
\textbf{You are tasked to classify a question into four types, following these guidelines:} \\
1. Output the type of the question based on its form of asking. Possible types are: \textit{yesno, list, factoid, summary}. \\
2. Just output one type without any descriptive information. \\
3. Summary questions are usually more general, but factoid questions are more specific. \\
4. You can infer the type according to the display forms of possible answers. \\
\textbf{Here are some examples:}\\
Question: Which DNA sequences are more prone to the formation of R-loops?\\
Output: list\\
Question: Are ultraconserved elements often transcribed?\\
Output: yesno\\
Question: What is clathrin?\\
Output: summary\\
Question: Which signaling pathway does sonidegib inhibit?\\
Output: factoid\\
\textbf{Please output the type of the following question:}\\
Question: <question>\\
Output: \\
\hline
\end{tabularx}
\caption{The prompt for question type classification.}
\label{prompt1}
\end{table}

\subsection{Prompts for Data Augmentation}
We perform an LLM-based QA pair augmentation task by providing the following prompt in Table \ref{prompt2} to GPT-4 and replacing <aim\_style> with the style label we desire to collect. For a particular question type, we also provide the answer style description at the place of <answer\_type> to restrict the generated text.
\begin{table}[H]
\centering
\begin{tabularx}{8cm}{|X|}
\hline
\textbf{You are tasked to answer the question with <aim\_style> language, following these guidelines:}\\
1. You can refer to the provided examples to learn the differences between professional and non-professional answers.\\
2. You can refer to the original <style> answer and rephrase into a different <aim\_style> answer.\\
3. For a <type> question, the <aim\_style> answer usually <answer\_style>.\\

\textbf{Here are examples of professional and non-professional answers:}\\

Question: What is gingipain?\\
Professional answer: Porphyromonas gingivalis is a keystone periodontal pathogen that has been associated with autoimmune disorders. The cell surface proteases Lys-gingipain (Kgp) and Arg-gingipains (RgpA and RgpB) are major virulence factors, and their proteolytic activity is enhanced by small peptides such as glycylglycine (GlyGly).\\

Question: Are reduced-nicotine cigarettes effective for smoking cessation?\\
Non-professional answer: Yes, reduced-nicotine cigarettes are effective for smoking cessation.\\

\textbf{Please give a <aim\_style> answer for the following question:}\\
Question: <question>\\
Original <style> answer: <original\_answer>\\
Output: \\
\hline
\end{tabularx}
\caption{The prompt for QA pairs generation.}
\label{prompt2}
\end{table}

\subsection{Prompts for Reasoning Step Calculation}
In our evaluation stage, we calculate the reasoning step count with the help of GPT-4 by reorganizing the answers into a step-by-step format and then giving the total step number at the end. The reorganization prompt is shown in Table \ref{prompt3}.
\begin{table}[H]
\centering
\begin{tabularx}{8cm}{|X|}
\hline
\textbf{You are an assistant to explain the reasoning path of the answer. Here are some requirements:}\\
1. Explain the reasoning path of the answer step by step with the content in both question and answer.\\
2. Provide the total steps at the last line, with the format: Total steps: <number>.\\
\textbf{Here are the question and the answer:} \\
Question: <question>\\
Answer: <answer>\\
\hline
\end{tabularx}
\caption{The prompt for reasoning step reorganization.}
\label{prompt3}
\end{table}

\section{More Implementation Details} \label{details}
\subsection{LLM-Augmented Type Classification} \label{typecls}
The type of question is classified into one of the four categories by GPT-4. However, there are some problems while performing the process. The major problem is the confusion between the meaning of types. One confusion occurs between factoid and summary, as they have similar sentence structures, such as the beginning of \textit{What is}. The difference is that summary questions are usually more general, such as \textit{What is Synucleinopathy?} and \textit{What is a zoonotic virus?}. However, factoid questions are more specific and aim to obtain a particular aspect of an entity, such as \textit{What is the function of a viral peplomer?}. Another confusion is between the list and factoid questions, which also have similar expressions but have different formats of answers. To address the above problems, as shown in Appendix A.1, we provide guidelines in our instruction to describe the distinctive information to help GPT4 better understand the differences.

\subsection{LLM-Augmented Data Balancing} \label{dataaugment}
As our PubMedPro dataset is constructed from academic QA scenarios, there are far more professional QA pairs than non-professional ones. To balance the number of QA pairs in each style, we perform an LLM-augmented data generation using the prompt shown in Table \ref{prompt2}. What needs to be emphasized is that we have tried different types of style descriptions to guide GPT4 to generate or rephrase our desired answers. One type of description follows our basic instruction format, such as \textit{explains the reason with detailed steps using technical professional expressions} for a professional question. Another type of description follows our type-based instruction format, such as \textit{has a list of items and explains each item with reasons in detailed steps using technical professional expressions} for a professional and list-type question. We evaluate the output of these two types of prompt for data generation and rephrasing by random sampling and manual checking and select the type-based description as the final version for data augmentation since it facilitates generation closer to reference answers.  

\subsection{Indicator Threshold Analysis} \label{threshold}
\begin{figure}[htbp]
    \begin{center}
    \includegraphics[width=0.48\textwidth]{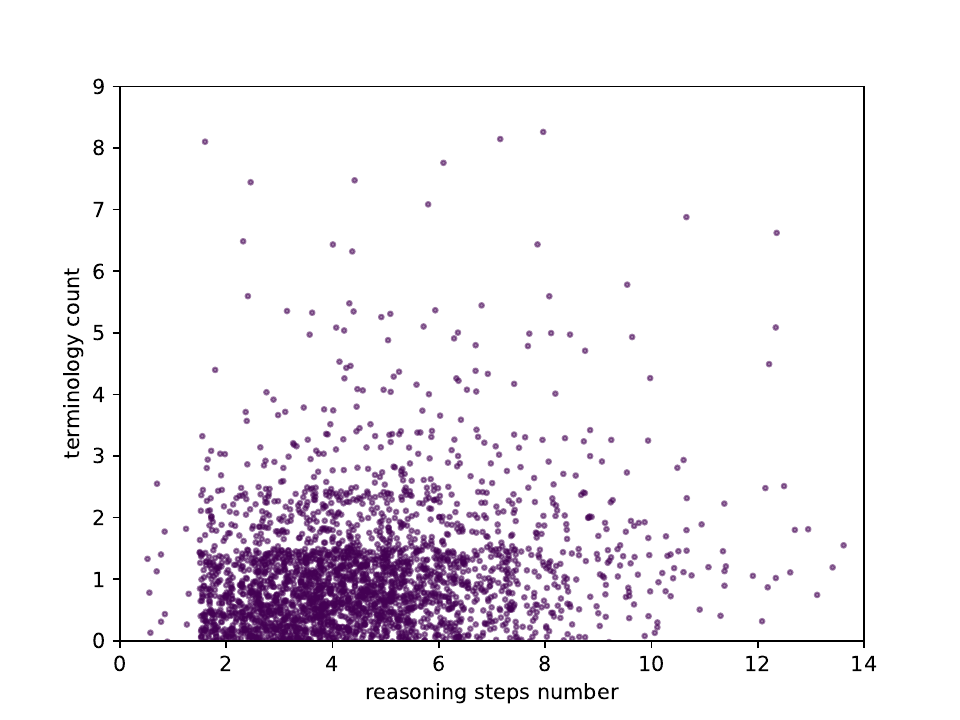} 
    \caption{Distribution of terminology count and reasoning step count from a part of PubMedPro dataset. Each value is added with a small random number for visual differentiation.}
    \label{distribution}
    \end{center}
\end{figure}

We explore for the thresholds of professional indicators according to human-labeled samples for the test data of binary classification. We recruit 3 volunteers to manually label 300 randomly selected QA pairs and screen out consistent labeling answers with three labels (professional, non-professional, unsure). We then adjust the number of terminology counts and reasoning steps of these answers that can distinguish professional and non-professional answers in order to fit the labels generated by humans. A distribution visualization of the two indicators is shown in Figure \ref{distribution}, demonstrating a larger number of reasoning steps than the terminology count should be specified to identify professional responses.

\subsection{PubMedPro Dataset Statistics}
The number of QA pairs at various stages for the two public datasets (BioASQ and PubMedQA) that constitute our PubMedPro dataset is shown in Table \ref{stats} below. As each question in BioASQ may contain multiple answers, we split them into multiple QA pairs.

\begin{table}[htbp]
\centering
\begin{tabular}{l|cc}
\bottomrule
\multirow{1}{*}{\textbf{Stage}}  & \multicolumn{1}{c}{\textbf{BioASQ}} & \multicolumn{1}{c}{\textbf{PubMedQA}}
\\ \hline      
\multirow{1}{*}{Raw} & \multicolumn{1}{c}{8,021} & \multicolumn{1}{c}{1,000}
\\ \hline
\multirow{1}{*}{} & \multicolumn{1}{c}{list: 1277} & \multicolumn{1}{c}{list: 160}
\\
\multirow{1}{*}{Typed} & \multicolumn{1}{c}{yes/no: 2150} & \multicolumn{1}{c}{yes/no: 359}
\\
\multirow{1}{*}{} & \multicolumn{1}{c}{summary: 1840} & \multicolumn{1}{c}{summary: 198}
\\
\multirow{1}{*}{} & \multicolumn{1}{c}{factoid: 2754} & \multicolumn{1}{c}{factoid: 283}
\\ \hline
\multirow{1}{*}{Balanced} & \multicolumn{1}{c}{20855} & \multicolumn{1}{c}{3145}
\\ \bottomrule
\end{tabular}
\caption{The statistical information of two original datasets that compose PubMedPro in various stages.}
\label{stats}
\end{table}

\subsection{Result Variation Caused by LLM} \label{variation}
There are two reasons that can cause unwanted performance variations in our task. The first comes from the formulation of prompts, which affects generating professional and non-professional answers the most. To avoid model-specific generation bias, we remove all predefined system prompts in open-source language models while fine-tuning and inference. To avoid competitors from generating too short or too long answers, we test extensive examples and add restrictive information related to the model. Another reason that may lead to variational results is from multiple requests, which affects both answer generation and reasoning step decomposition. We run each model three times and calculate the average values of the indicators or the number of steps as usable results. We also set appropriate hyperparameters such as Top-K, Top-P, and Temperature for each model during generation.

\subsection{Human Evaluation Details} \label{humaneval}
We recruit 5 volunteers for the style switching and language fluency evaluation of our generated text from postgraduate students at our college. We provide the instruction as follows: \textit{You are asked to rate answers generated by an LLM for text generation research in a domain QA scenario. Please evaluate the English answers provided in terms of their professional distinctiveness and language fluency as follows: 1. For the given pair of English answers, evaluate their professional differentiation; the better the distinction in professionalism, the higher the score (1-5 points). 2. For each English answer provided, rate the fluency of the language; the easier it is to read, the higher the score (1-5 points).} We keep the rating results that are completed by the annotators and calculate the average scores and success rates for each model, as discussed in the experiment section. In addition, we invite two researchers in the medical domain to evaluate the accuracy of the results, simply by asking: \textit{You are asked to determine whether the answers generated by an LLM are correct. Please provide TRUE, FALSE, or UNKNOWN for each answer.} The metric is calculated as the proportion of results deemed correct out of all results.

\section{Case Study} \label{casestudy}
In this section, we provide examples to highlight the improvement of our model after applying our tuning method. The examples in Table \ref{performance4} showcase how ProSwitch and our base model Llama2-Chat generate answers to a question with a different number of technical terms. ProSwitch is apt to generate more domain terms compared to the baseline model when providing a professional response. Table \ref{performance5} shows that ProSwitch tends to deliver more logically linked concise sentences than Llama2-Chat while answering a yes/no question in professional style. Table \ref{performance6} shows that ProSwitch can generate an answer with more domain terms and a better sentence structure in the meantime than Llama2-Chat.
\begin{table*}[!ht]
\centering
\begin{tabularx}{16cm}{X|X}
\hline
\multicolumn{2}{c}{\textbf{Question:} What is the biological role of Neddylation?} \\
\hline
\textbf{Professional Answer} & \textbf{Non-Professional Answer} \\
\hline
\textbf{ProSwitch}: The biological role of Neddylation is to regulate the activity of \underline{enzymes} involved in ... This process involves the \underline{covalent attachment} of a small protein called \underline{NEDD8} to the target enzymes ... The NEDD8 protein is a member of the \underline{ubiquitin-like} protein family and ... The Neddylation pathway is a highly regulated process that plays a crucial role in maintaining \underline{cellular homeostasis} and ... & \textbf{ProSwitch}: Neddylation is when proteins get a special thing called a Nedd8 tag. This tag helps the proteins do their job and also helps them get broken down. It's important for the cell to know when to break down proteins and when to keep them around. \\
\textbf{Llama2-Chat}: Neddylation is a \underline{post-translational modification} that plays a crucial role in various cellular processes, including protein function, localization, and degradation. It's a way for proteins to be modified after they're made, which can affect how they work and how long they last. The biological role of Neddylation can be broken down into several key functions ... & \textbf{Llama2-Chat}: Neddylation is like a molecular glue that helps proteins do their thing in the body. It adds a little tag to proteins that help them interact with other proteins and stuff. Like, it's like a secret handshake for proteins! \\
\hline
\end{tabularx}
\caption{An example to illustrate the differentiation of technical terms (highlighted with underlines) in professional and non-professional answers generated by our ProSwitch model and Llama2-Chat model. Some content is omitted due to space constraints.}
\label{performance4}
\end{table*}

\begin{table*}[!ht]
\centering
\begin{tabularx}{16cm}{X|X}
\hline
\multicolumn{2}{c}{\textbf{Question:} Can Levetiracetam cause disturbance in sleeping?} \\
\hline
\textbf{Professional Answer} & \textbf{Non-Professional Answer} \\
\hline
\textbf{ProSwitch}: Yes, Levetiracetam can cause disturbance in sleeping.$\mathbin\Vert$ This is due to the drug's ability to affect the activity of the neurotransmitter GABA,$\mathbin\Vert$ which plays a crucial role in regulating sleep.$\mathbin\Vert$ Levetiracetam can increase the activity of GABA,$\mathbin\Vert$ leading to a decrease in the amount of time spent in slow-wave sleep and rapid eye movement (REM) sleep.$\mathbin\Vert$ This can result in a decrease in the quality of sleep and an increase in daytime sleepiness.$\mathbin\Vert$ Additionally, Levetiracetam can also cause vivid dreams and nightmares,$\mathbin\Vert$ which can further disrupt ... & \textbf{ProSwitch}: Yes, Levetiracetam can cause disturbance in sleeping.$\mathbin\Vert$ It can make you feel tired and sleepy during the day, $\mathbin\Vert$ but it can also make it harder to fall asleep at night. \\
\textbf{Llama2-Chat}: Levetiracetam, a commonly used anticonvulsant medication, can potentially cause disturbances in sleep.$\mathbin\Vert$ Here's why: ... $\mathbin\Vert$ While it is primarily used to treat epilepsy, it has also been shown to have anxiolytic and sedative effects in some individuals.$\mathbin\Vert$ However, these effects can sometimes lead to disturbances in sleep patterns. & \textbf{Llama2-Chat}: Levetiracetam is a medicine that helps control seizures in people with epilepsy.$\mathbin\Vert$ But, it can also make some people feel sleepy or have trouble sleeping.$\mathbin\Vert$ This is because it affects the way the brain sends signals that help us fall asleep and stay asleep.$\mathbin\Vert$ It's like when you're trying to ... \\
\hline
\end{tabularx}
\caption{An example to illustrate the differentiation of reasoning steps (separated by double vertical lines) in professional and non-professional answers generated by our ProSwitch model and Llama2-Chat model. Some content is omitted due to space constraints.}
\label{performance5}
\end{table*}

\begin{table*}[!ht]
\centering
\begin{tabularx}{16cm}{X|X}
\hline
\multicolumn{2}{c}{\textbf{Question:} Are there transposon-free regions in mammalian genomes?} \\
\hline
\textbf{Professional Answer} & \textbf{Non-Professional Answer} \\
\hline
\textbf{ProSwitch}: Yes, there are transposon-free regions in mammalian genomes. These regions are known as transposon-free regions (\underline{TFRs}) and are found in both the human and mouse genomes. \uwave{Despite} the presence of over 3 million transposons ... The majority of human TFRs are found to be similar to their \underline{orthologous} counterparts in the mouse, indicating that ... \uwave{Additionally}, many human TFRs also overlap with orthologous TFRs in the marsupial opossum, suggesting that these regions have been conserved across different \underline{mammalian lineages} ... & \textbf{ProSwitch}: Yeah, some parts of mammalian genomes don't have transposons. These regions are called "transposon-free regions" or "TFRs". They're like special zones in the genome where transposons don't hang out. \\
\textbf{Llama2-Chat}: Yes, there are transposon-free regions in mammalian genomes. These regions are known as "gap regions" or "gap-like sequences" and are found in the vicinity of transposon insertion sites. \uwave{To explain further}, transposons are mobile genetic elements that can insert themselves into new locations within a genome. When a transposon inserts itself into a region of the genome that is already occupied by a gene or regulatory element, it can disrupt the function of that gene or element ... & \textbf{Llama2-Chat}: Yeah, some regions in mammalian genomes are free of transposons. These regions are usually found near the centromere or at the ends of chromosomes. \\
\hline
\end{tabularx}
\caption{An example to illustrate the differentiation of technical terms (highlighted with underlines) and conjunctive adverbs (highlighted with wave underlines) for logical structure in professional and non-professional answers generated by our ProSwitch model and Llama2-Chat model. Some content is omitted due to space constraints.}
\label{performance6}
\end{table*}


\end{document}